\documentclass{ecai} 

\usepackage{latexsym}
\usepackage{amssymb}
\usepackage{amsmath}
\usepackage{amsthm}
\usepackage{booktabs}
\usepackage{enumitem}
\usepackage{graphicx}
\usepackage{color}
\usepackage{svg}
\usepackage{algorithm}
\usepackage{algpseudocode}
\usepackage{xspace}
\usepackage{makecell}

\newtheorem{theorem}{Theorem}
\newtheorem{lemma}[theorem]{Lemma}

\newcommand{\BibTeX}{B\kern-.05em{\sc i\kern-.025em b}\kern-.08em\TeX}

\begin{document}


\begin{frontmatter}




\title{FedMP: Tackling Medical Feature Heterogeneity in Federated Learning from a Manifold Perspective}


\author[A]{\fnms{Zhekai}~\snm{Zhou}}
\author[A]{\fnms{Shudong}~\snm{Liu}}
\author[B]{\fnms{Zhaokun}~\snm{Zhou}}
\author[B]{\fnms{Yang}~\snm{Liu}}
\author[B]{\fnms{Qiang}~\snm{Yang}}
\author[A]{\fnms{Yuesheng}~\snm{Zhu}}
\author[A]{\fnms{Guibo}~\snm{Luo}}

\address[A]{Peking University}
\address[B]{the Hong Kong Polytechnic University}


\begin{abstract}
Federated learning (FL) is a decentralized machine learning paradigm in which multiple clients collaboratively train a shared model without sharing their local private data. However, real-world applications of FL frequently encounter challenges arising from the non-identically and independently distributed (non-IID) local datasets across participating clients, which is particularly pronounced in the field of medical imaging, where shifts in image feature distributions significantly hinder the global model's convergence and performance. To address this challenge, we propose FedMP, a novel method designed to enhance FL under non-IID scenarios. FedMP employs stochastic feature manifold completion to enrich the training space of individual client classifiers, and leverages class-prototypes to guide the alignment of feature manifolds across clients within semantically consistent subspaces, facilitating the construction of more distinct decision boundaries. We validate the effectiveness of FedMP on multiple medical imaging datasets, including those with real-world multi-center distributions, as well as on a multi-domain natural image dataset. The experimental results demonstrate that FedMP outperforms existing FL algorithms. Additionally, we analyze the impact of manifold dimensionality, communication efficiency, and privacy implications of feature exposure in our method.
\end{abstract}

\end{frontmatter}


\section{Introduction}
As the need for privacy-preserving machine learning grows, federated learning (FL) has emerged as a promising paradigm for decentralized model training. FL enables collaborative model training by exchanging only model parameters between clients and a central server, eliminating the need to share raw data. 
However, real-world applications of FL often face significant challenges caused by data heterogeneity across clients. In most cases, local datasets are non-independent and identically distributed (non-IID), typically manifesting in the following two forms\cite{kairouz2021advances}: (1) label distribution skew, where the label space varies across clients, such as in face recognition tasks where certain identities appear only within specific devices; and (2) feature distribution skew, where the underlying data characteristics differ significantly across clients, e.g., in handwritten digit recognition tasks where users exhibit highly distinct writing styles. These non-IID conditions can lead to client drift\cite{karimireddy2020scaffold} during local training under the classic FedAvg\cite{mcmahan2017communication} framework, which negatively impacts the convergence and performance of the global model.

\begin{figure}[h]
\centering
\includegraphics[width=.43\textwidth]{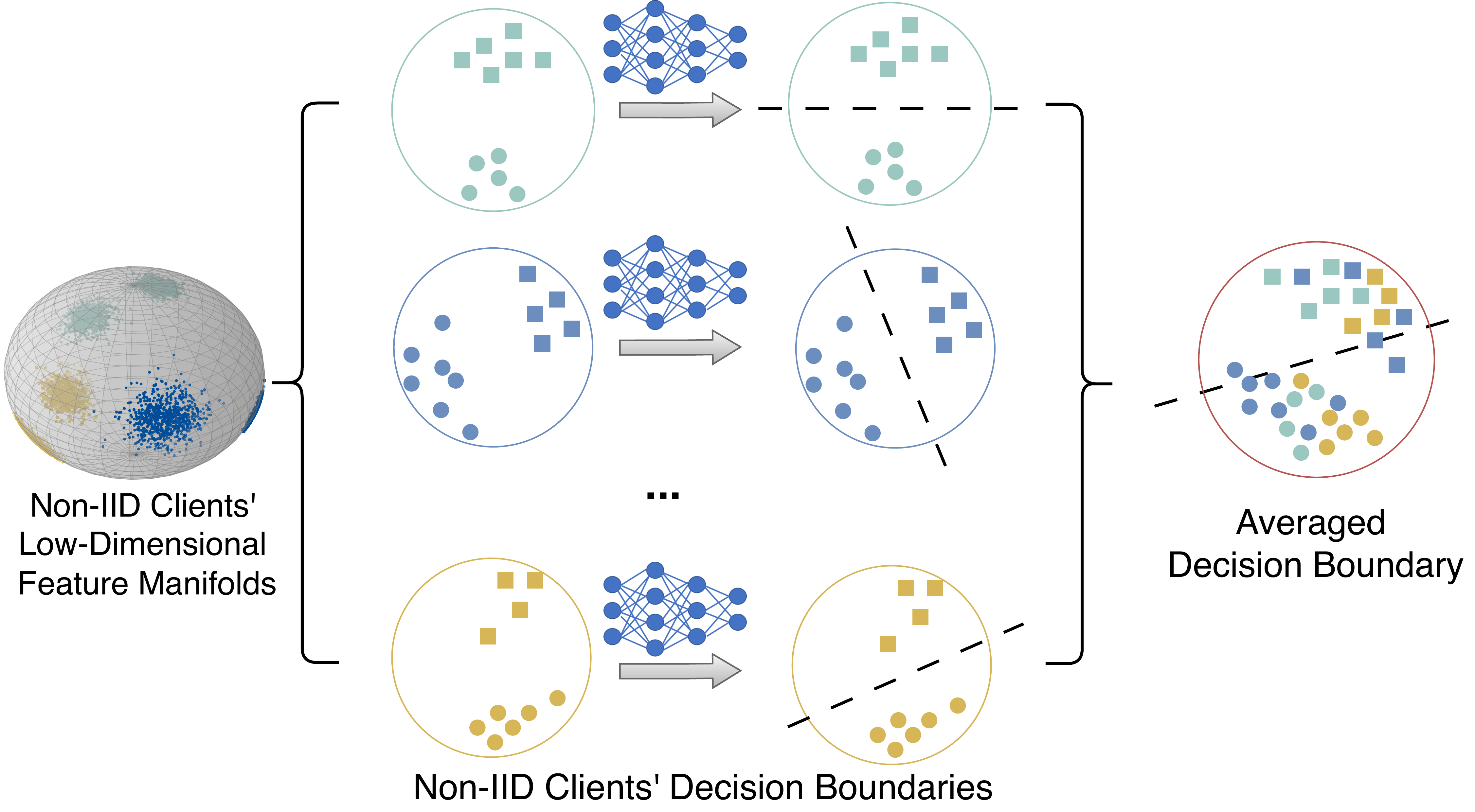}
\vspace{-8pt}
\caption{Adverse impact of feature skew on global model aggregation.}
\label{fig:feature_shift}
\end{figure}

Numerous FL algorithms have been proposed to address the non-IID problem. However, most representative approaches\cite{li2020federated,wang2020tackling,karimireddy2020scaffold,acar2021federated,li2021fedrs,tan2022fedproto,gao2022feddc,mu2023fedproc,fan2024federated} focus primarily on mitigating label distribution skew, and are evaluated in experiments that typically simulate non-IID settings within a single dataset, assigning samples of different labels using Dirichlet distributions or adopting pathological non-IID partitioning schemes\cite{mcmahan2017communication} in which each client has access to only a subset of class labels.
However, these methods are usually less effective in handling feature distribution skew, which is highly prevalent in practical FL scenarios, especially in medical imaging analysis. For example, hospitals in different regions may collect diagnostic records for the same disease, yet their imaging data are acquired using different types of medical devices. Variations in imaging hardware and acquisition protocols can lead to differences in image intensity and contrast, ultimately causing heterogeneity in feature distributions\cite{dou2019domain}.
The adverse impact of feature distribution skew on model average aggregation is illustrated in Figure~\ref{fig:feature_shift}, where samples from different clients and categories are represented by different colors and shapes, respectively. 
In addressing feature space non-IID challenges, many existing FL algorithms have only been evaluated on a limited number of multi-domain natural image datasets\cite{li2021fedbn,chen2023fraug,zhang2023fedala}, such as Office-Home\cite{venkateswara2017deep}, with a noticeable lack of experimental validation on medical imaging datasets, where more severe feature space non-IID is commonly observed. 
Moreover, they are constrained to the perspective of improving feature consistency\cite{li2021model,zhou2023fedfa}.
Other approaches relying on feature augmentation or pseudo-sample generation often suffer from instability in practical medical applications due to scarce and heterogeneous training data. Existing generative models, even after fine-tuning, often struggle to produce high-quality medical images, while training them from scratch demands significant client-side computational resources and incurs substantial time costs\cite{zhang2025non}.

\begin{figure}[ht]
\centering
\vspace{-6pt}
\includegraphics[width=\columnwidth]{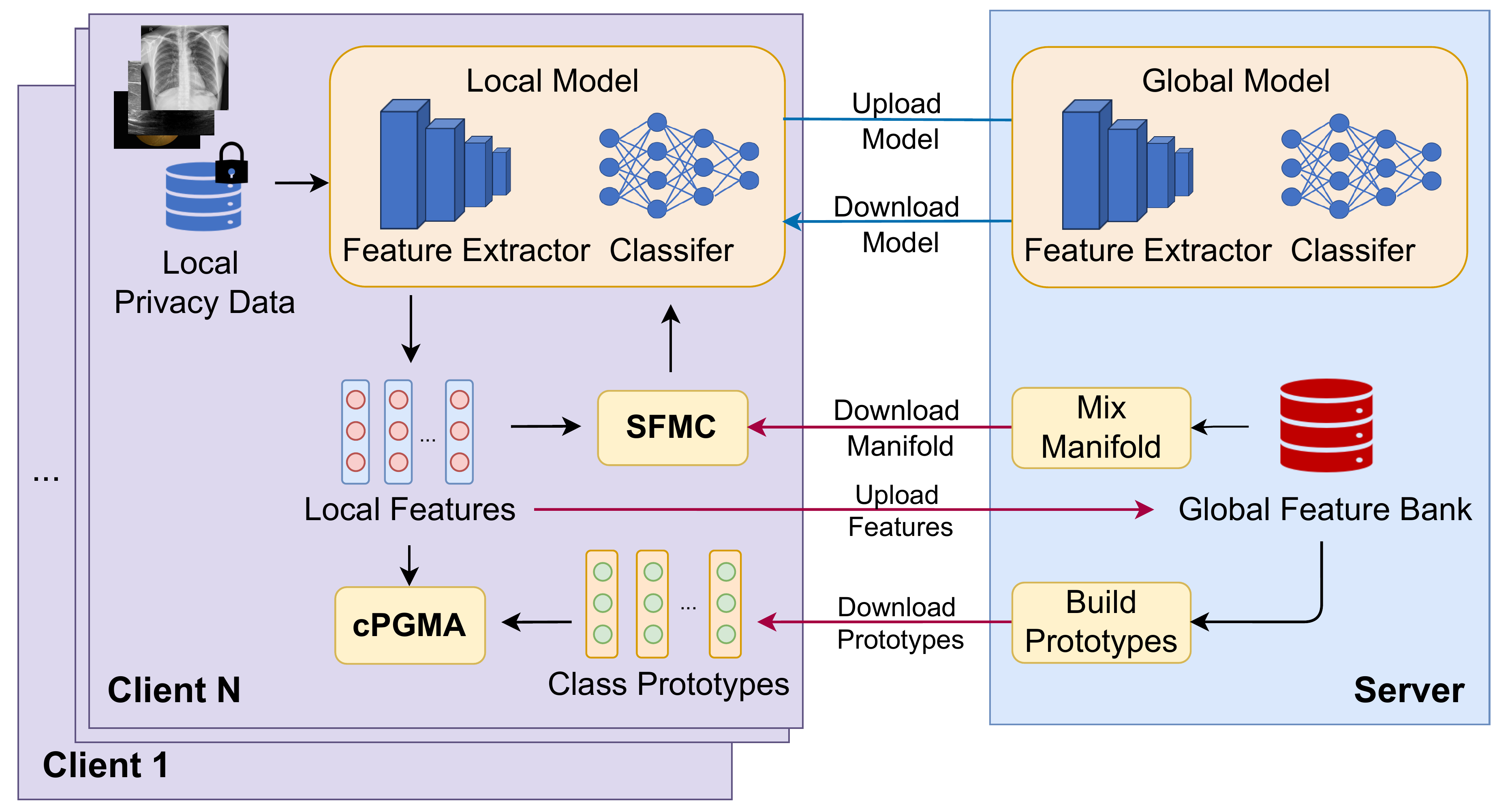}
\vspace{-18pt}
\caption{The framework of our proposed method FedMP.}
\label{fig:arch}
\vspace{-5pt}
\end{figure}

In the context of manifold learning\cite{meilua2024manifold,melas2020mathematical}, high-dimensional data samples with the same semantic label are typically distributed along a shared low-dimensional manifold. However, under non-IID feature space conditions, due to the shift in the appearance characteristics of same-category samples across clients, semantically similar samples will be mapped onto disjoint low-dimensional sub-regions. This results in a fragmented and incomplete global manifold structure, which hinders the learning of consistent decision boundaries and ultimately degrades the generalization ability and cross-client classification performance of the global model, as shown in Figure~\ref{fig:feature_shift}.
To overcome this issue, we propose a novel and effective algorithm from the perspective of structure completion and geometric alignment of low-dimensional manifolds.
It also avoids the training and transmission of generative models as well as the transmission of synthetic data, thus reducing computational demands, time cost, and communication overhead compared to the latest FL methods based on generative models.
The overall architecture is illustrated in Figure~\ref{fig:arch}. 
Our main contributions are summarized as follows.
\begin{itemize}
    \item We propose the stochastic feature manifold completion technique, which reconstructs and completes latent manifolds from partial observations across clients, in order to alleviate the impact of non-IID feature spaces on classifier.

    \item We propose the class-prototype guided manifold alignment technique, which aligns class-specific feature manifolds using shared prototypes to enhance cross-client consistency.

    \item We integrate the above techniques into a unified FL framework, FedMP, which, in our comprehensive experiments, outperforms state-of-the-art FL algorithms on multiple medical and natural image benchmarks, including real-world feature non-IID datasets.
    
    \item We extend FedMP to a few-shot FL setting, demonstrating competitive global accuracy with significantly reduced communication overhead.
    

\end{itemize}

\section{Related Work}
The classic FedAvg\cite{mcmahan2017communication} aggregates local models after multi-epoch updates, but non-IID clients’ data significantly hinder the global model performance and generalization. To address this issue, current FL research has focused mainly on the following directions.

\subsection{FL Algorithms via Model Update Calibration}
In order to enforce the consistency between global and local models, FedProx\cite{li2020federated} adds a regularization term in loss function to constrain local updates. 
SCAFFOLD\cite{karimireddy2020scaffold} employs control variables to reduce divergence. FedNova\cite{wang2020tackling} introduces an aggregation rule that accounts for the number of local update steps. FedDyn\cite{acar2021federated} incorporates a dynamic regularization term based on the global model. 
FedRS \cite{li2021fedrs} applies a restricted softmax to local classes to enhance discriminative performance.
FedDC \cite{gao2022feddc} explicitly aligns the client-server update differences using local drift variables and gradient correction. 
Elastic Aggregation\cite{chen2023elastic} emphasizes that parameters less sensitive to the variation of model output can be updated more freely, while minimizing updates to more sensitive parameters.
These methods are mainly effective under label non-IID settings, where they primarily reduce the optimization drift without reconciling feature space discrepancies.
Although later methods\cite{chen2023best, chen2022personalized} employ knowledge distillation to smooth updates under feature non-IID conditions, they may compromise the performance of global model.

\subsection{FL Algorithms via Feature Space Optimization}
Researchers also have proposed methods in feature representation levels. FedBN\cite{li2021fedbn} emphasizes handling local feature heterogeneity by personalizing BN layers on each client. FedFA\cite{zhou2023fedfa} uses global feature anchors to jointly align feature spaces and calibrate classifiers. MOON \cite{li2021model} applies contrastive learning to enforce feature similarity between global and local models, reducing model drift. FedPAC\cite{xu2023personalized} aligns local features with global feature centers and introduces dynamic classifier collaboration. FedUFO\cite{zhang2021federated} aligns client-specific feature spaces through adversarial learning.
FedMR\cite{fan2024federated} performs manifold reshaping locally, including preventing intra-class feature collapse and calibrating feature spaces using class prototypes. 
These methods primarily focus on enhancing feature consistency across clients, but do not consider building a more complete feature space to better reduce the bias of the aggregated classifier.

\subsection{FL Algorithms via Data Augmentation}
Some latest FL methods tackle the feature non-IID issue through data augmentation.
FedAlign\cite{gupta2025fedalign} applies local feature augmentation via MixStyle.
FRAug\cite{chen2023fraug} performs personalized data augmentation through feature generation on the client side to improve global model adaptability across domains. DENSE\cite{zhang2022dense} uses a set of client models as discriminators to train a generator that produces pseudo-samples for model aggregation. In addition, various approaches based on diffusion models\cite{ho2020denoising,zhang2023federated,yang2023exploring,yang2023one} have been proposed to augment training data within FL frameworks on multi-center datasets.
However, FL utilizing generative approaches incurs significant time and computational costs and is dependent on the quality of the generator.

In contrast to the approaches mentioned above, our proposed FL algorithm, based on the reconstruction and adjustment of client local manifolds, provides a more straightforward, effective, and resource-saving way to mitigate feature heterogeneity and improve model performance under complex data distributions.

\section{Method}
\subsection{Problem Statement}
Our method is designed to address the challenge of clients’ feature non-IID data in FL systems, which can be formally defined as follows:
Assume there are $N$ clients participating in the FL process. Each client $i \in \{1,2, \ldots, N\}$ holds a local dataset $D_i$ of size $M_i$, with a feature space $\mathcal{U}_i \subset \mathbb{R}^d$ and a label space $\mathcal{Y}_i \subset \mathbb{N}$. The label space is assumed to be IID across clients, i.e., $\mathcal{Y}_1, \mathcal{Y}_2, \ldots, \mathcal{Y}_N \overset{\text{i.i.d.}}{\sim} \mathcal{Y}$, while the feature distributions are heterogeneous such that $\mathcal{U}_i, \mathcal{U}_j \not\overset{\text{i.i.d.}}{\sim} \mathcal{U}, \forall i \ne j,\ i, j \in \{1, 2, \dots, N\}$.
In the FedAvg framework, the local model on each client is updated using multiple rounds of stochastic gradient descent (SGD) on its respective datasets before being uploaded and averaged to reduce the frequency of communication between clients and the server\cite{mcmahan2017communication,lin2018don}. However, the discrepancy among $\mathcal{U}_i$ across clients leads to local model drift. Specifically, local models tend to overfit their own data distributions and task objectives, which results in significant deviation of the server-aggregated model parameters from the global optimum in the sample space $D = \bigcup_{i=1}^{N} D_i$, negatively impacting both the convergence speed and the eventual performance of the global model\cite{li2020federated,karimireddy2020scaffold,luo2023influence}.

\subsection{Motivation}
In high-dimensional spaces, semantically similar data samples are often mapped to nearby regions in a lower-dimensional latent space. However, as previously discussed, due to factors such as device heterogeneity or sampling bias, client-local datasets in practical FL scenarios typically exhibit substantial feature distribution heterogeneity. Inspired by the idea of manifold learning, we model the feature set of each client as a collection of class-conditional low-dimensional manifolds embedded in a shared latent space\cite{melas2020mathematical}, as illustrated in Figure~\ref{fig:feature_shift}. In feature non-IID settings, these manifold structures encounter two major challenges: (1) 
Due to data sparsity or distributional bias within individual clients, intra-class features may only cover partial regions of the underlying manifold, resulting in fragmented geometric structures. 
(2) Manifold substructures corresponding to the same class from different clients often exhibit significant discrepancies in geometric properties such as orientation, scale, and density. These inconsistencies increase the difficulty of aggregating data representations across clients and hinder the ability of the global model to generalize discriminative patterns in the latent space.


\subsection{Proposed Method}
To address the aforementioned problems, we propose a new FL optimization framework, FedMP, grounded in the perspective of manifold modeling. The framework is composed of two synergistic modules. (1) Stochastic feature manifold completion (SFMC): During local training, external embeddings are stochastically introduced to augment the client's feature manifold. This enhances the geometric completeness and representational capacity of intra-class manifolds, particularly under sparse or biased local distributions. (2) Class-prototype guided manifold alignment (cPGMA): A set of global class prototypes is constructed to serve as geometric anchors in the latent space. These prototypes guide the alignment of class-conditional manifold structures across clients, promoting geometric consistency within each class, and facilitating the global feature aggregation. 

Similar to FedAvg, our framework consists of two main procedures: (1) \emph{Server Update}: The central server collects model weights and auxiliary data uploaded by clients, performs aggregation, and redistributes the updated model to all participants. (2) \emph{Client Update}: Each client receives the updated model parameters and other data from the server and performs local optimization using its private dataset.
Unlike FedAvg, FedMP decomposes each client's local classification neural network into two components: a feature extractor $f(x; \theta_i^f)$, typically implemented using a ResNet\cite{he2016deep} backbone, and an MLP classifier $h(x; \theta_i^c)$, where a sample is first embedded in a local feature space $\mathcal{U}_i$, then mapped to the label space $\mathcal{Y}_i$. 
Instead of training the entire model using only local raw data, FedMP leverages feature embeddings shared across multiple clients to fine-tune the local classifier. This enables the FL system to reconstruct a more complete low-dimensional manifold structure for each class across heterogeneous client feature spaces $\{\mathcal{U}_i\}_{i=1}^N$, and to train client local classifier directly over the mixed distribution of them. 
Moreover, FedMP aims to align the manifold structures across clients by minimizing the Hausdorff distance between sub-manifolds corresponding to the same semantic class but originating from different clients. Through federated training, this encourages feature distributions from different clients to become more consistent with an IID-like global structure. The overall loss function for client $i$ is formulated as Eq.(\ref{eq:2}),
where the first term $\ell_i^{\text{local}}$ is the standard cross-entropy loss, and $\ell_i^{\text{SFMC}}$ and $\ell_i^{\text{cPGMA}}$ correspond to the optimization objectives of the two modules in FedMP, respectively. We adopt a self-adaptive weighting strategy for overall loss. The terms $(\cdot)^*$ indicate that the gradients are detached during backpropagation.
This design allows the two auxiliary losses to be adaptively balanced relative to the primary task loss, while maintaining stable training dynamics. 
\vspace{-12pt}\begin{eqnarray}\label{eq:2}
\mathcal{L}_i = \ell_i^{\text{local}} + \frac{\ell_i^{\text{SFMC}}}{(\ell_i^{\text{SFMC}}/\ell_i^{\text{local}})^*}  + \frac{\ell_i^{\text{cPGMA}}}{(\ell_i^{\text{cPGMA}}/\ell_i^{\text{local}})^*}
\end{eqnarray}
In the following subsections, we provide a detailed explanation of these two modules. 

\subsubsection{Stochastic Feature Manifold Completion}
\begin{figure}[h]
\centering
\includegraphics[width=.95\columnwidth]{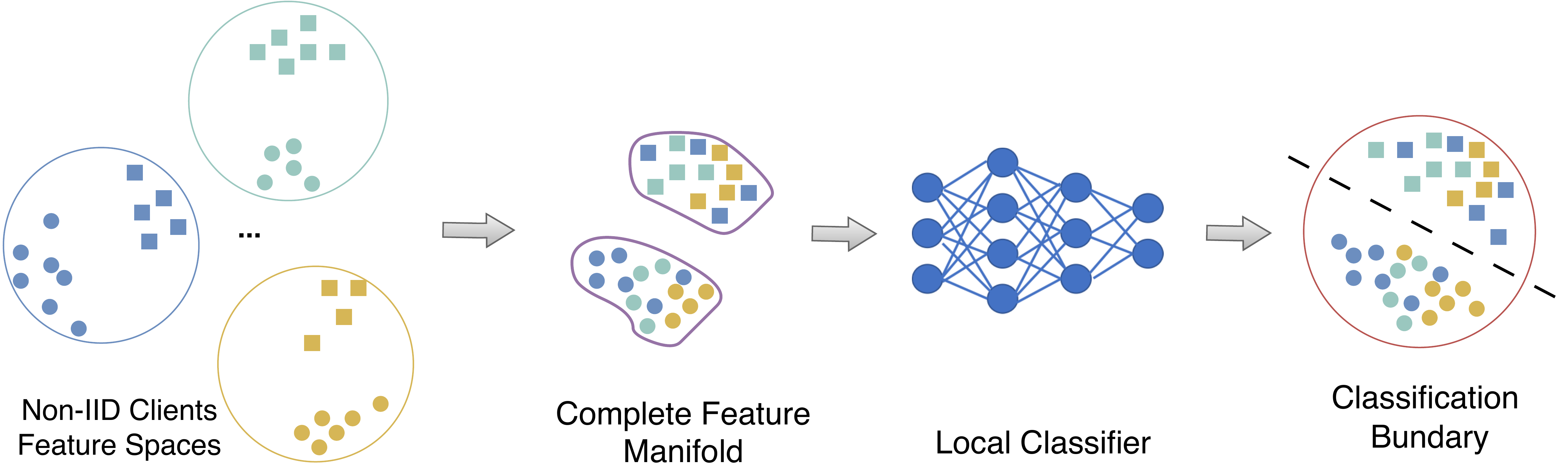}
\vspace{-6pt}
\caption{Role of SFMC in feature heterogeneity scenario.}
\label{fig:SFMC}
\end{figure}

SFMC module constructs an extended and mixed low-dimensional manifold by combining the embeddings of a client's local data with randomly sampled embeddings from other clients. By training the local classifier on a completed manifold structure, the model gains improved discriminative capability across diverse feature domains, thus reducing the phenomenon of client drift, as shown in Figure~\ref{fig:SFMC}.

We provide a formal and mathematical description as follows. 
Each client $i$ holds a local dataset $\mathcal{D}_i = \{(x_{i,j}, y_{i,j})\}_{j=1}^{M_i}$, where $x_{i,j} \in \mathbb{R}^{D_0}$ denotes the input data, and $y_{i,j} \in \{0, 1, \dots, K-1\}$ is the corresponding class label. 
Let the client’s feature extractor be defined as $\mathcal{F}_i : \mathbb{R}^{D_0} \to \mathbb{R}^d, \mathcal{F}_i(x) := f(x; \theta_i^f)$, which maps raw inputs into a d-dimensional latent space. 
For each class $c$, the class-conditional feature manifold on client $i$ is defined as $\mathcal{M}_i^{(c)} = \{ \mathcal{F}_i(x_{i,j}) \mid y_{i,j} = c \} \subset \mathbb{R}^d.$ Due to the non-IID settings, each local manifold $\mathcal{M}_i^{(c)}$ only captures a fragment of the global class manifold, i.e., 
$\mathcal{M}_i^{(c)} \subsetneq \mathcal{M}^{(c)} \text{, where } \mathcal{M}^{(c)} = \bigcup_{i=1}^{N} \mathcal{M}_i^{(c)}.$
As a result, the local classifier $\mathcal{H}_i: \mathbb{R}^d \to \mathbb{R}^K, \mathcal{H}_i(x) := h(x; \theta_i^c)$ is optimized based only on a partial, possibly fragmented manifold, i.e., 
\begin{equation}
\mathcal{H}_i^* = \arg\min_\mathcal{H} \mathbb{E}_{c \sim \mathcal{Y}}\left[\mathbb{E}_{u \sim \mathcal{M}_i^{(c)}}\left[ \ell_{\text{CE}}(\mathcal{H}(u), c)\right]\right],
\end{equation}
which may lead to suboptimal or biased class boundaries in the global feature space.
SFMC module reconstructs the local manifold by incorporating feature embeddings sampled from other clients: $\hat{\mathcal{M}}_i^{(c)} \subset \mathcal{M}^{(c)} \setminus \mathcal{M}_i^{(c)}$, and forms an extended feature manifold $\tilde{\mathcal{M}}_i^{(c)} = \mathcal{M}_i^{(c)} \cup \hat{\mathcal{M}}_i^{(c)}, \text{where } \mathcal{M}_i^{(c)} \subset \tilde{\mathcal{M}}_i^{(c)} \subset \mathcal{M}^{(c)}$. The local classifier is then trained on this more complete manifold structure, as 
\begin{equation}
\tilde{\mathcal{H}}_i^* = \arg\min_\mathcal{H} \mathbb{E}_{c \sim \mathcal{Y}}\left[\mathbb{E}_{u \sim \tilde{\mathcal{M}}_i^{(c)}} \left[ \ell_{\text{CE}}(\mathcal{H}(u), c)\right]\right].
\end{equation}
It can be asserted that $d_H(\tilde{\mathcal{M}}_i^{(c)}, \mathcal{M}^{(c)}) < d_H(\mathcal{M}_i^{(c)}, \mathcal{M}^{(c)})$, where $d_H$ denotes the Hausdorff distance between two manifolds, as
\begin{equation}
\begin{array}{l}
d_H(\mathcal{M}_p, \mathcal{M}_q) \\
\displaystyle
= \max \left\{ 
\sup_{a \in \mathcal{M}_p} \inf_{b \in \mathcal{M}_q} \|a - b\|,\;
\sup_{b \in \mathcal{M}_p} \inf_{a \in \mathcal{M}_q} \|a - b\|
\right\}.
\end{array}
\end{equation}
We theoretically derive that under such conditions, the local classifier can learn decision boundaries that are more aligned with the global optimum, thus improving generalization and cross-client consistency, which is clarified by the following Lemma~\ref{lemma1}. The complete proof of it is summarized in the supplementary material.

\begin{lemma}
\label{lemma1}
Let $\mathcal{M} = \{ \mathcal{M}^{(c)} \}_{c=0}^{K-1}$ denote the global class-conditional feature manifold. 
Suppose that client classifier \( \mathcal{H}_i \) is trained on the local manifold \( \mathcal{M}_i = \{ \mathcal{M}^{(c)}_i \}_{c=0}^{K-1} \), and classifier \( \mathcal{H}_j \) is trained on \( \mathcal{M}_j = \{ \mathcal{M}^{(c)}_j \}_{c=0}^{K-1} \). 
If the Hausdorff distance between the local and global manifolds satisfies $d_H(\mathcal{M}_i, \mathcal{M}) < d_H(\mathcal{M}_j, \mathcal{M}),$
then the classifier $\mathcal{H}_i$ is expected to converge to a solution closer to the global optimum $\mathcal{H}^*$, compared to $\mathcal{H}_j$.
\end{lemma}

SFMC module is implemented in two main steps: (1) During the final epoch of local gradient descent training, each client extracts intermediate representations of its private data from a selected layer of the feature extractor (e.g., one of the convolutional layers of the ResNet backbone). These intermediate features are flattened, labeled with their corresponding category, and then uploaded to the server together with the locally trained parameters of both the feature extractor and classifier. 
(2) The server stores the embeddings received from multiple clients in the global feature bank. To reduce communication overhead, it constructs a mixed manifold by randomly sampling embeddings from the global feature bank and distributes it to individual clients.
The external embeddings are restored to the same dimensionality by the client as if they had participated in local model computation and are mixed with the client's local embeddings to form a more complete class-conditional manifold. The classifier is then trained on this set of mixed features to optimize the classification objective. The local optimization at client $i$ can be formulated as 
\begin{equation}\label{sfmcloss}
\ell_i^{\text{SFMC}} = \sum_{k=1,\, k \ne i}^N \sum_{j \in \mathcal{I}_k} \ell_{\text{CE}} (h(u_{k,j}; \theta_i^c), y_{k,j}),\end{equation}
where $\mathcal{I}_k$ is the set of indices of sampled embeddings of client $k$.

\subsubsection{Class-Prototype Guided Manifold Alignment}
To mitigate the client feature shift caused by multiple epochs of local training on non-IID data, where the feature extractor gradually overfits the local distribution and drifts away from a globally consistent representation, we introduce a manifold alignment module that involves collaborative optimization between the server and clients.

A naive solution to address feature skew is to directly align feature distributions across different clients to a common distribution. However, such an approach often degrades the discriminative capacity of the feature extractor\cite{zhao2020multi}. Instead, our method tackles the problem from a manifold perspective: We treat the local feature distributions on each client as labeled sub-manifolds in a low-dimensional space and aim to align them geometrically under the guidance of class-wise prototypes, which represent the mean embeddings of each class\cite{snell2017prototypical} and encode the semantic location of each class in the embedding space. We estimate global class prototypes over distributed clients in FL systems to serve as shared geometric anchors. Under this formulation, client-specific manifolds are encouraged to align around the same semantic centers, which in turn facilitates more consistent and generalizable decision boundaries across the global feature space.

\begin{figure}[h]
\centering
\vspace{-6pt}
\includegraphics[width=.82\columnwidth]{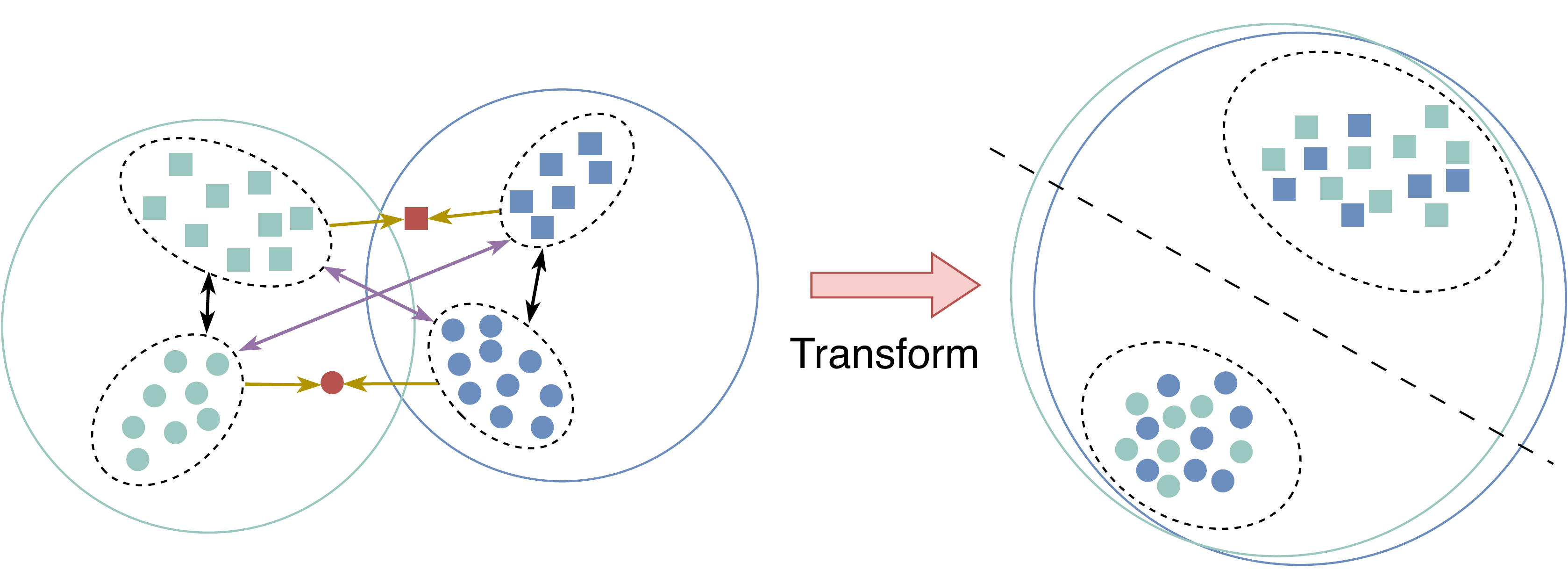}
\vspace{-8pt}
\caption{Impact of complete FedMP on non-IID feature spaces.}
\label{fig:alignment}
\end{figure}

The specific alignment procedure is as follows. 
(1) After the clients finish uploading their data, new feature embeddings for each class stored in the global feature bank are smoothed using an exponential moving average (EMA) to reduce instability caused by noisy updates from early or fluctuating model states. The server then performs weighted average aggregation across clients for each class to compute the global prototype, again using an EMA-based update. The result is a set of $K$ prototype vectors (one per class), representing the approximate global geometric centers of the semantic manifolds. These prototypes are then distributed back to clients. Assume $\mathcal{I}_c$ is the set of indices of samples belonging to class $c$ in a batch. The update process for the global prototype of class $c$ is shown in Eq.(\ref{p-update-eq:1}) and Eq.(\ref{p-update-eq:2}), where $\mu_{\text{client}}$ and $\mu_{\text{server}}$ are the momentum coefficients for local and global EMA, respectively.
\begin{equation}\label{p-update-eq:1}
\bar{z}_i^{(c)} \leftarrow (1-\mu_{\text{client}})\bar{z}_i^{(c)} + \frac{\mu_{\text{client}}}{|\mathcal{I}_c|} \sum_{j \in \mathcal{I}_c} \mathcal{F}_i(x_{i,j})
\end{equation}
\vspace{-6pt}  
\begin{equation}
\label{p-update-eq:2}
p^{(c)} \leftarrow (1 - \mu_{\text{server}}) p^{(c)} + \mu_{\text{server}} \sum_{i=1}^N \frac{M_i \cdot \bar{z}_i^{(c)}}{\sum_{i=1}^NM_i}
\end{equation}
(2) On each client, during mini-batch gradient descent, local feature extractors output embeddings for each sample. For each class $c$, a set of local embeddings is grouped, and the training is guided by encouraging these embeddings to move closer to the corresponding global prototype, which is implemented via a prototype alignment loss term, calculated as
\begin{equation}
\label{alignloss}
\ell_i^{\text{cPGMA}} = - \sum_{c=0}^{K-1} \frac{1}{|\mathcal{I}_c|} \sum_{j \in \mathcal{I}_c} (\frac{\mathcal{F}_i(x_{i,j})}{\| \mathcal{F}_i(x_{i,j}) \|_2})^\top \cdot (\frac{p^{(c)}}{\| p^{(c)} \|_2}).
\end{equation}
We can derive the following Lemma~\ref{lemma2}, with a detailed proof provided in the supplementary material. By combining Lemma~\ref{lemma1} and Lemma~\ref{lemma2}, we finally derive a theoretical justification for the effectiveness of cPGMA module in enhancing the performance of the global model in FL under non-IID conditions.
\begin{lemma}
\label{lemma2}
Let \( \mathcal{M} \) denote the global class-conditional feature manifold, and let \( \mathcal{M}_i^{(t)} \) be the class-conditional feature manifold of client \( i \) at communication round \( t \). Suppose that the client locally optimizes a loss function as (\ref{alignloss}). Then the Hausdorff distance between the local and global manifolds satisfies $ d_H(\mathcal{M}_i^{(t+1)}, \mathcal{M}) < d_H(\mathcal{M}_i^{(t)}, \mathcal{M})$.
\end{lemma}
The formalized pseudo-code of the complete training process is shown in Algorithm~\ref{alg:fedmp}.
Figure~\ref{fig:alignment} provides an intuitive illustration of how FedMP behaves in a non-IID feature space scenario. 
Consider a simplified case with two clients participating in federated training. The features extracted from each client's local data are visualized using green and blue points, respectively. Each client possesses two classes of samples, denoted by squares and circles. 
In FedAvg, local models learn to distinguish classes based solely on their own incomplete local feature manifolds, as indicated by the black arrows in the figure. Additionally, in FedMP, through cPGMA module, local feature manifolds are pulled toward the global geometric centers of each class, achieving the calibration of the client's class-specific manifold structure, as shown by the yellow arrows. 
Simultaneously, SFMC module enables feature-level data augmentation across clients. 
By integrating embeddings sampled from other clients, it completes the local manifold structure, facilitating a single client’s classification model to better capture classification objectives in heterogeneous feature spaces of other clients, as shown by the purple arrows. 
As a result of these collaborative mechanisms, FedMP gradually adapts local feature distributions, enabling clients' classifiers to operate in a more consistent and globally coherent feature space.

\vspace{-4pt}
\begin{minipage}{.95\columnwidth}
\begin{algorithm}[H]
\caption{FedMP}
\label{alg:fedmp}
\begin{algorithmic}[1]

\State \textbf{Input:} Communication rounds $T$, client number $N$, local epochs $E$, datasets $\{\mathcal{D}_i\}_{i=1}^N$, class number $K$, hyperparameters $\mu_{\text{client}}, \mu_{\text{server}}$.
\State \textbf{Initialize:} Server model $\theta_s$, class prototypes $p^{(c)}=0\ ,\forall c \in \{0,1,\ldots,K-1\}$; for each client $i$, initialize $\theta_i=[\theta_i^f,\theta_i^c]$.
\For{$t = 1$ to $T$}
    \For{each client $i$ \textbf{in parallel}}
        \State $(\theta_i, U_i) \gets \textsc{ClientUpdate}(i, \theta_s)$
    \EndFor
    \For{each class $c$ \textbf{in parallel}}
        \State Update all clients' manifold centers as Eq.(\ref{p-update-eq:1})
        \State Update global class prototypes as Eq.(\ref{p-update-eq:2})
    \EndFor
    \State $\theta_s \gets \sum_{i=1}^{N}\frac{|\mathcal{D}_i|}{\sum_{i=1}^N |\mathcal{D}_i|}\theta_i$
\EndFor
\State \Return $\theta_s$
\Function{ClientUpdate}{$i$, $\theta_s$}
    \State Download $\theta_s$, $\{U_j\}_{j\ne i}$ and $\{p^{(c)}\}_{c=0}^{K-1}$
    \State $\theta_i \gets \theta_s$
    \For{$e = 1$ to $E$}
        \For{each mini-batch $(X,y) \sim \mathcal{D}_i$}
            \State $u \gets f(X;\theta_i^f)$, $\hat{y} \gets h(u;\theta_i^c)$
            \State $\ell_i^{\text{local}} \gets \text{CrossEntropy}(\hat{y},y)$
            \State Calculate $\mathcal{L}_i$ as Eq.(\ref{sfmcloss}), Eq.(\ref{alignloss}), and Eq.(\ref{eq:2})
            \State Update $\theta_i$ with SGD on $\mathcal{L}_i$
        \EndFor
    \EndFor
    \State Collect batch features: $U_i \gets \{(u,y)\}$
    \State \Return $(\theta_i, U_i)$
\EndFunction
\end{algorithmic}
\end{algorithm}
\end{minipage}

\subsection{Communication-Efficient Few-Shot FedMP}
As described in the previous section, FedMP involves the exchange of model parameters and feature data between clients and the server, which leads to substantial network communication overhead and computational burden on the server. Therefore, based on the FedMP optimization strategy, we propose a communication-efficient few-shot FL framework.

In this framework, each client first trains its local model on its own dataset for multiple rounds. Once the local features extracted by the feature extractor become stable, a one-time communication with the server is triggered: (1) Each client uploads its locally trained model for a single round of model aggregation. (2) Each client uploads multiple batches of current embeddings to the server. (3) The server performs feature exchange and distribution, and computes class prototypes by aggregating features of the same category across clients at once. These global class prototypes are then sent back to the clients to guide manifold alignment. Subsequently, each client enters the second stage of local training, where: (1) Feature embeddings from local data and received cross-client features are combined to form a more complete manifold structure. (2) The classifier is trained over this reconstructed manifold through multiple epochs of gradient descent. (3) Simultaneously, the global class prototypes are used as geometric anchors to align the client's feature manifolds, thus guiding the training of the feature extractor. After multiple rounds of local training in this second stage, the communication process between the server and clients can be repeated to allow the local model to gain more global knowledge. Finally, when the local training on each client is completed, a final communication is performed. All clients upload their local models to the server, which can ensemble the predictions of these models. By decreasing the frequency of model aggregation and prototype updates, few-shot FedMP significantly reduces the communication overhead in the FL system.


\section{Experiments}
We conduct extensive experiments to validate the effectiveness of FedMP. We evaluate the method on five medical imaging classification datasets, two of which are real-world federated datasets collected from multiple sources. We also test FedMP on a natural image dataset with domain shift to further demonstrate its generalization capability. We further analyze the convergence speed, latent feature space dimensionality, and privacy leakage risks, and perform ablation studies to assess the contribution of each component in FedMP.

\begin{table*}[htbp]
  \centering
  \caption{Accuracy comparison on multiple datasets (mean ± std).}
  \label{tab:results}
  \vspace{-6pt}
  \begin{tabular}{lccccccc}
    \toprule
    & TB & DR & COVID & Breast & NeoJaundice & Office-Caltech10 \\
    \midrule
    Centralized & 92.83$\pm$0.74 & 81.49$\pm$0.26 & 96.74$\pm$0.19 & 90.65$\pm$0.30 & 83.38$\pm$0.13 & 98.78$\pm$0.25 \\
    \hline
    Single  & 69.75$\pm$0.74 & 62.06$\pm$0.58 & 93.56$\pm$0.40 & 73.25$\pm$0.91 & 71.82$\pm$0.50 & 93.63$\pm$0.42 \\
    FedAvg  & 84.45$\pm$0.42 & 70.55$\pm$0.38 & 94.79$\pm$0.20 & 84.29$\pm$0.30 & 80.71$\pm$0.13 & 97.31$\pm$0.09 \\
    FedProx & 85.84$\pm$0.25 & 72.11$\pm$0.35 & 95.53$\pm$0.25 & 83.65$\pm$0.30 & 81.51$\pm$0.13 & 97.18$\pm$0.33 \\
    MOON    & 84.45$\pm$0.42 & 72.14$\pm$0.62 & 95.95$\pm$0.04 & 84.08$\pm$0.52 & 82.22$\pm$0.45 & 97.51$\pm$0.10 \\
    FRAug   & 86.70$\pm$0.64 & 73.12$\pm$0.37 & 95.78$\pm$0.05 & 85.35$\pm$0.52 & 81.78$\pm$0.13 & 97.70$\pm$0.09 \\
    FedBN   & 81.86$\pm$0.42 & 72.35$\pm$0.33 & 95.87$\pm$0.10 & 84.71$\pm$0.52 & 82.14$\pm$0.38 & 97.52$\pm$0.19 \\
    Elastic Aggregation & 86.53$\pm$0.73 & 72.16$\pm$0.61 & 95.81$\pm$0.13 & 83.23$\pm$0.79 & 80.71$\pm$0.55 & 97.84$\pm$0.37 \\
    FedMR   & 85.41$\pm$0.42 & 73.15$\pm$0.17 & \textbf{96.09$\pm$0.14} & 86.20$\pm$0.60 & 81.24$\pm$0.12 & 97.45$\pm$0.16 \\
    FedMP (Ours) & \textbf{88.08$\pm$0.42} & \textbf{75.96$\pm$0.31} & 95.78$\pm$0.09 & \textbf{88.75$\pm$0.30} & \textbf{82.49$\pm$0.13} & \textbf{98.04$\pm$0.32} \\
    \bottomrule
  \end{tabular}
\end{table*}

\subsection{Experiments Setup}
\textbf{Datasets.} We first use three common medical imaging classification datasets: (1) NeoJaundice\cite{wang2023real}, a binary classification dataset of neonatal skin photographs for diagnosing jaundice; (2) COVID-QU-Ex\cite{tahir2021covid}, a chest X-ray dataset categorized in COVID-19, non-COVID-19 infections, and normal; and (3) Breast\cite{al2020dataset}, a breast ultrasound dataset categorized in normal, benign, and malignant.
Additionally, we combine two real-world multi-center medical imaging datasets: (1) DR, a diabetic retinopathy dataset consisting of fundus images from three independent medical institutions (APTOS 2019 Blindness Detection\cite{aptos2019-blindness-detection}, Retino\cite{wang2023real}, and IDRID\cite{h25w98-18}), classified into five severity levels; and (2) TB\cite{rahman2020reliable}, a binary classification dataset for the diagnosis of tuberculosis, composed of chest X-ray images from three geographically distinct sources (India, Shenzhen, and Montgomery).
We also include Office-Caltech10\cite{gong2012geodesic}, a natural image dataset with feature skew, which contains four visually distinct domains: Amazon, Caltech, DSLR, and Webcam.
For non-IID datasets TB, DR, and Office-Caltech10, the training set of each domain-specific subset is assigned to a single client to simulate realistic federated heterogeneity, while the test sets are merged for global evaluation. For other datasets, we randomly split the training data into partitions and distribute them evenly among five clients. 

\textbf{Model.} In all experiments, we adopt ResNet-50 as the backbone of the feature extractor, which consists of four multi-bottleneck stages. The output of each stage in ResNet-50 is considered as a candidate for intermediate features in our method. Consequently, the earlier stages of the ResNet-50 network are designated as the feature extractor in our FedMP framework, while the latter stages combined with MLP form the classifier.

\textbf{Baselines.} We compare FedMP against nine baseline methods: 
(1) Centralized, i.e., collecting all data from clients for centralized training, resulting in privacy leakage; 
(2) Single, i.e., training a separate model on each client and performing one-time model averaging; 
(3) FedAvg\cite{mcmahan2017communication}, the basic FL algorithm; (4) FedProx\cite{li2020federated} and (5) Elastic Aggregation\cite{chen2023elastic}, both of which are effective in addressing label heterogeneity; 
(6) FedBN\cite{li2021fedbn}, classic FL algorithm designed to address feature non-IID challenges; methods leveraging feature alignment or augmentation, including (7) MOON\cite{li2021model}, (8) FRAug\cite{chen2023fraug}, and (9) FedMR\cite{fan2024federated}. We carefully select the coefficient of these baselines and report their best results in our experiments. Detailed hyperparameter settings are documented in the supplementary material.

\textbf{Parameters.} For all experiments, the initial learning rate of each client model is set to $10^{-4}$, using the Adam optimizer with hyperparameters $\beta_1 = 0.9, \beta_2 = 0.999$, and a weight decay of $5 \times 10^{-4}$. 
Both training and inference are performed with a batch size of $64$. 
Our method is configured with hyperparameters $\mu_{\text{client}} = 0.5$ and $\mu_{\text{server}} = 0.7$ in all comparison experiments.

\subsection{Experimental Results and Analysis}
\subsubsection{FL Performance with Multi-round Communication}
Under the setting where multiple rounds of model and feature data transfer are performed, our method achieves the best performance in five datasets, as shown in Table~\ref{tab:results}. Each experiment is repeated with three random seeds, and we report the mean and standard deviation of the test accuracy of the global model on the three runs. In particular, FedMP demonstrates superior performance on two real-world multi-center medical imaging datasets, with improvements of 3.6\% (TB) and 5.4\% (DR) compared to FedAvg, and over 1.0\% improvement compared to the best-performing baselines (FRAug and FedMR). These results demonstrate that FedMP performs well in realistic FL scenarios. Furthermore, on the Breast dataset, FedMP also achieves a significant improvement of approximately 4.5\% over FedAvg and a gain of over 1.0\% compared to the best-performing baseline, FedMR. The results on the Office-Caltech10 dataset also show the robustness of our federated method to natural image tasks under multi-domain distribution.

\begin{figure}[H]
    \centering
    \begin{minipage}{0.2\textwidth}
        \centering
        \includegraphics[width=\linewidth]{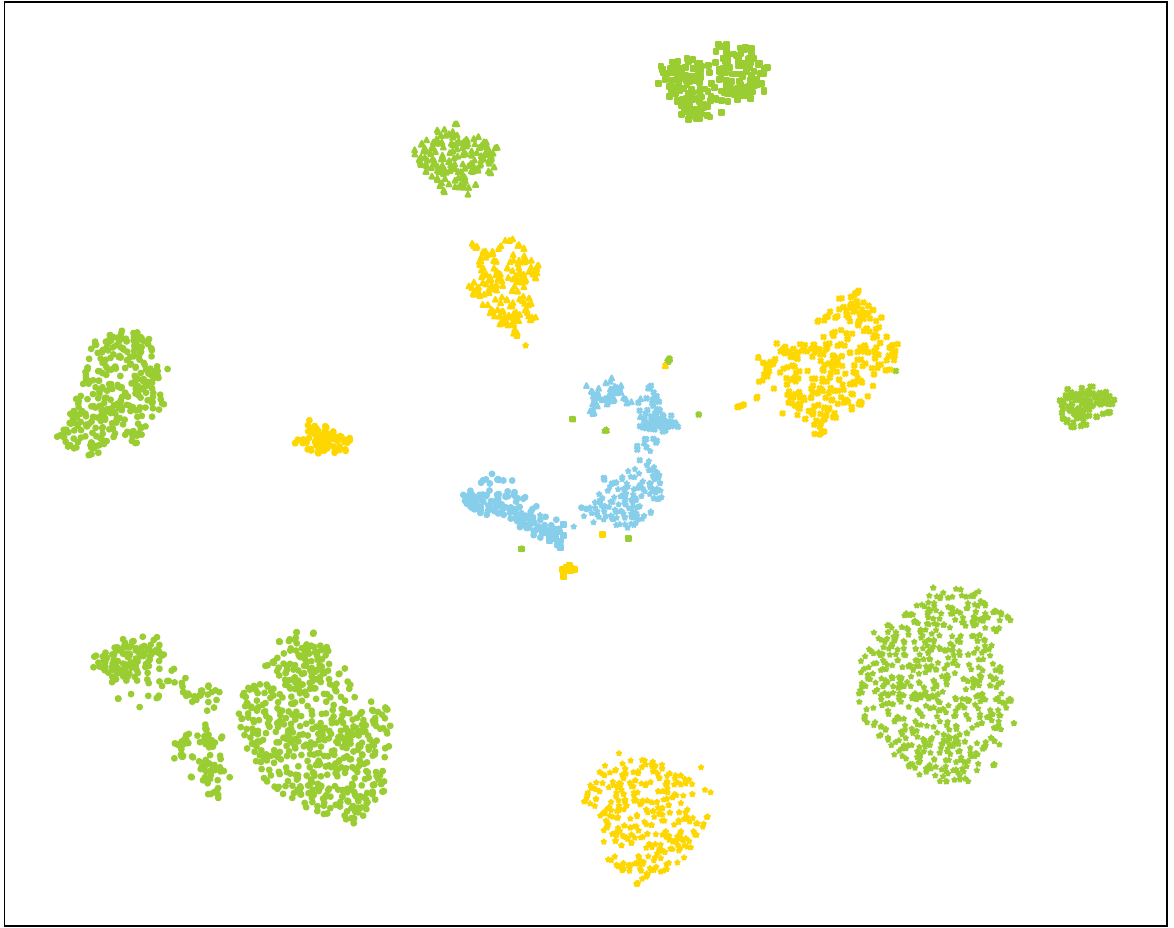}
    \end{minipage}
    \hspace{0.05\textwidth}
    \begin{minipage}{0.2\textwidth}
        \centering
        \includegraphics[width=\linewidth]{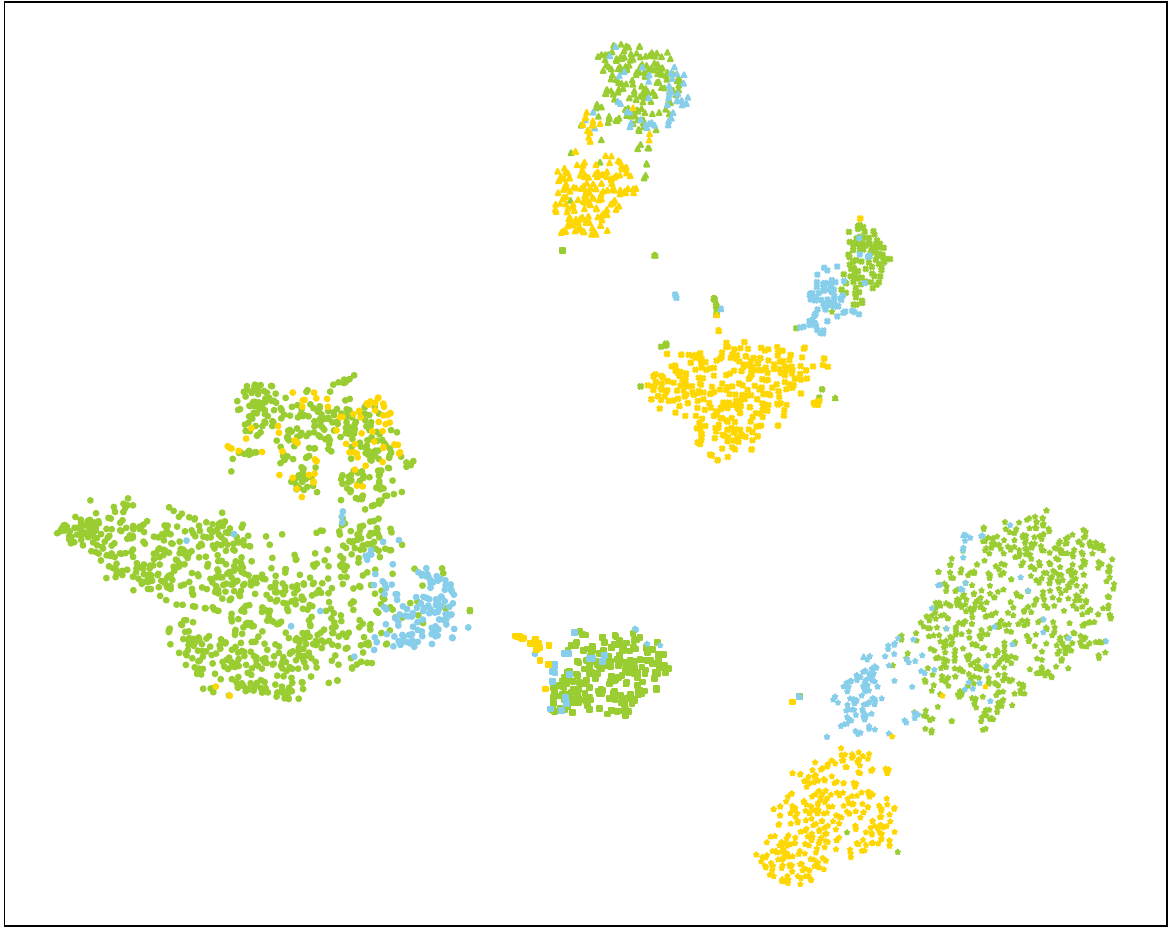}
    \end{minipage}
    \caption{T-SNE visualization (DR) for FedAvg (left) and FedMP (right).}
    \label{fig:diff}
\end{figure}

We employ the t-SNE technique to visualize the outputs of the global feature extractor on heterogeneous client data under both the FedAvg and FedMP algorithms. As shown in Figure~\ref{fig:diff}, different shapes represent different classes (five in total), and different colors indicate samples from different non-IID clients (three in total). It can be observed that after federated training with FedMP, the feature distributions of client data become more aligned and closer to an IID-like configuration, which facilitates the global classifier to learn more consistent and effective classification boundaries across heterogeneous client data, thereby achieving improved accuracy.

\subsubsection{Impact of Different Manifold Dimensions}
We investigate the impact of performing FedMP under different latent manifold dimensions on model performance and training process.
In each experiment, we extract embeddings from a specific stage of ResNet-50 for communication and perform optimization strategies in a matched spatial dimension. 

Our experiments show that using higher-dimensional manifolds built from shallower network layers for optimization leads to instability in early-stage training due to less stable prototypes and more complex manifold structure. 
It also increases communication, memory, and computation overhead in FL system, while resulting in improved classification accuracy after convergence. Therefore, a trade-off must be considered between costs and potential gains in model performance. In our setup, using embeddings from the third or fourth stage achieves a favorable balance, matching baseline convergence rounds while delivering higher accuracy, as summarized in Table~\ref{tab:feature_layer_comparison}. We provide a detailed comparison in the supplementary material.

\begin{table}[ht]
\vspace{-4pt}
\centering
\caption{Convergence rounds and accuracy of different dimensions used.}
\vspace{-6pt}
\begin{tabular}{l|r|c|c}
\hline
\textbf{Methods} & \textbf{Dimensions} & \makecell{\textbf{DR Accuracy} \\ \textbf{(\#Rounds)}} & \makecell{\textbf{TB Accuracy} \\ \textbf{(\#Rounds)}} \\
\hline
FedMP (1st stage) & 1,048,576 & 77.13 (170) & 89.12 (80) \\
FedMP (2nd stage) & 524,228   & 77.13 (150) & 88.60 (80) \\
FedMP (3rd stage) & 262,144   & 76.65 (110) & 88.60 (70) \\
FedMP (4th stage) & 2,048     & 76.25 (95) & 88.08 (60) \\
FedAvg                   & -         & 70.20 (90)  & 84.45 (65) \\
\hline
\end{tabular}
\label{tab:feature_layer_comparison}
\vspace{-13pt}
\end{table}

\subsubsection{Ablation Study}
To demonstrate the effectiveness of each component in FedMP,
we perform ablation experiments. Specifically, 
we evaluate the classification performance of the global model when using only the SFMC or cPGMA module during federated training. As shown in Table~\ref{tab:ablation}, each module individually contributes to performance improvement, and together they provide complementary benefits, highlighting their synergistic role within FedMP framework.

\begin{figure}[H]
    \centering
    \begin{minipage}{0.2\textwidth}
        \centering
        \includegraphics[width=\linewidth]{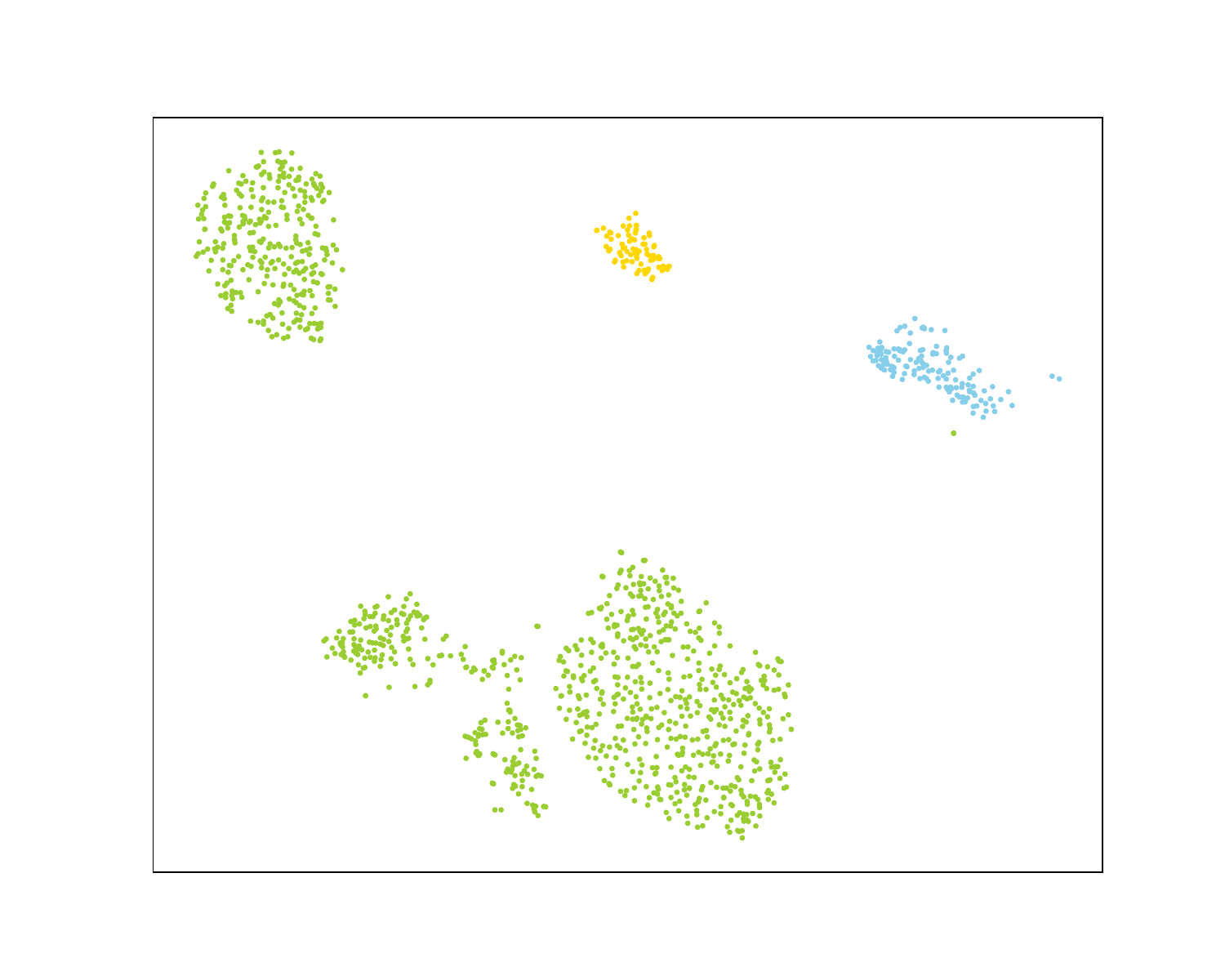}
    \end{minipage}
    \hspace{0.05\textwidth} 
    \begin{minipage}{0.2\textwidth}
        \centering
        \includegraphics[width=\linewidth]{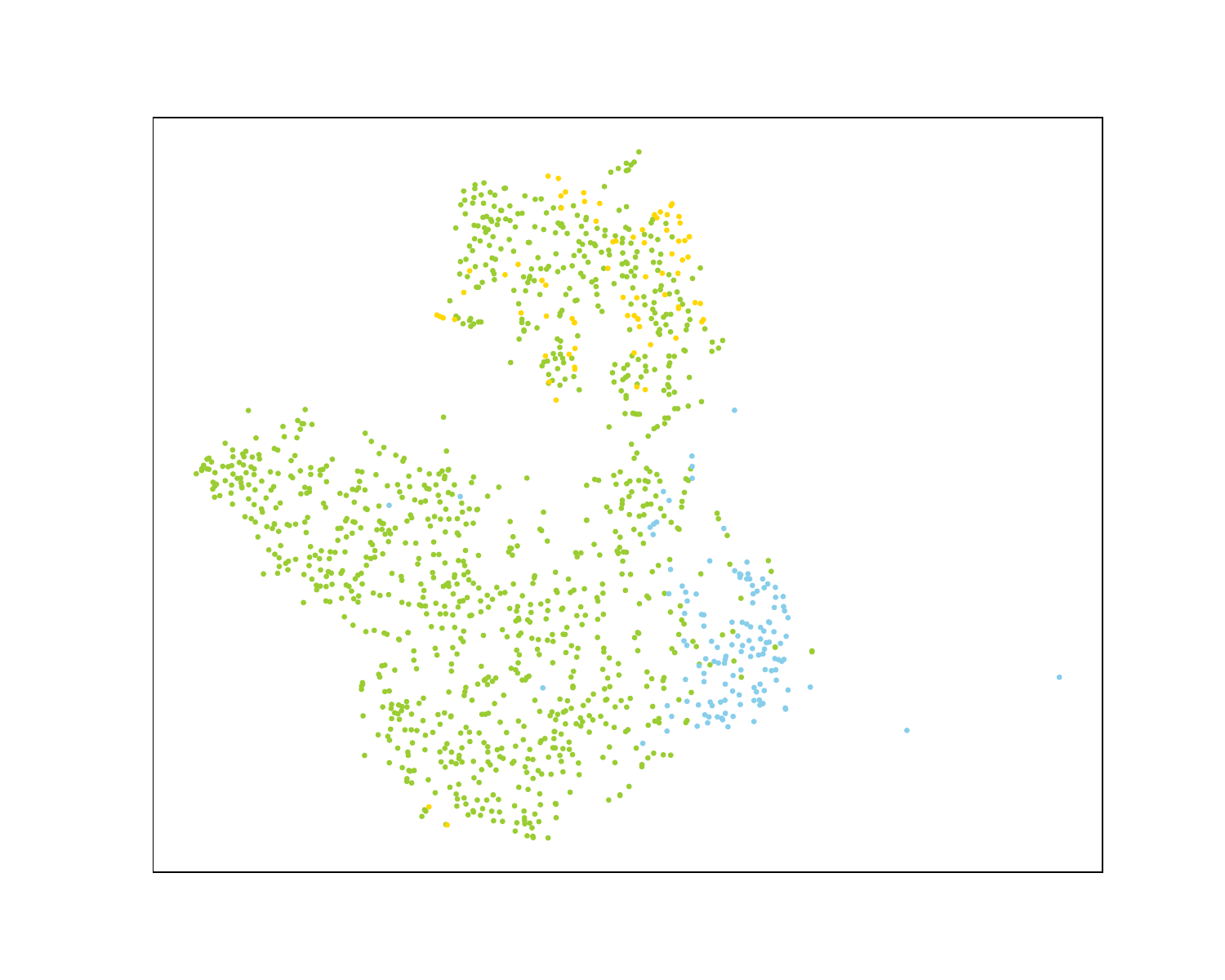}
    \end{minipage}
    \caption{T-SNE visualization for FedMP w/o and w/ cPGMA in one class.}
    \vspace{-12pt}
    \label{fig:diff1}
\end{figure}

We further illustrate this observation through feature visualizations of DR. 
When only the SFMC module is applied, although local classifiers are trained on the completed low-dimensional manifold, samples of the same class from different clients still exhibit shifted distributions in the feature extractor. Even within a single client's dataset, samples of the same class may cluster in separate regions, as shown on the left of Figure~\ref{fig:diff1}.
With the cPGMA module, the sub-manifolds of the same class from different clients are progressively pulled toward a shared geometric center, eventually aligning into a continuous manifold structure, as shown on the right of Figure~\ref{fig:diff1}.
\begin{table}[ht]
\vspace{-3pt}
\centering
\caption{Ablation experiment results (DR).}
\vspace{-6pt}
\begin{tabular}{l|c|c|c|c}
\hline
\textbf{Methods} & \makecell{No Optimization\\(FedAvg)} & \makecell{Only\\SFMC} & \makecell{Only\\cPGMA} & \makecell{SFMC+\\ cPGMA} \\
\hline
\textbf{Accuracy} & 70.20 & 74.10 & 72.59 & 75.61 \\
\hline
\end{tabular}
\label{tab:ablation}
\end{table}

\subsubsection{Communication Overhead Reduction}
\begin{figure}[h]
\centering
\includegraphics[width=.78\columnwidth]{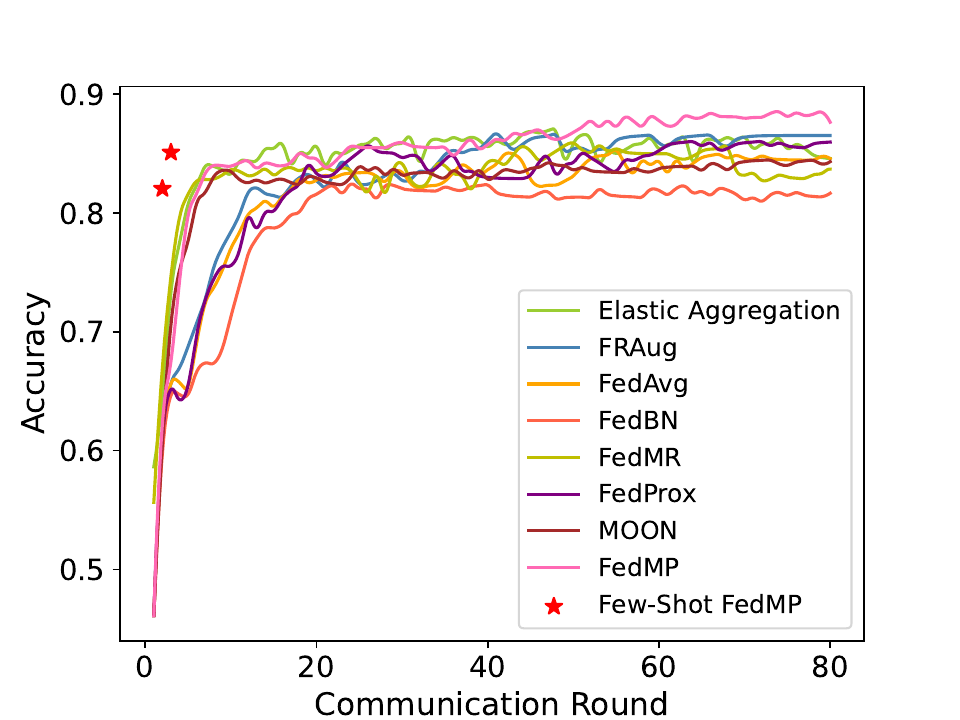}
\vspace{-8pt}
\caption{Accuracy curves across different FL methods (TB).}
\label{fig:TB_Curve}
\end{figure}

We measure the detailed communication cost of various FL methods on TB dataset, as shown in Table~\ref{tab:communication_efficiency} and Figure~\ref{fig:TB_Curve}. FedMP can be flexibly applied to different scenarios. When aiming for optimal model performance, multi-round FedMP introduces additional communication overhead due to the transmission of feature vectors; however, the overhead remains lower than that of FRAug or other FL methods based on diffusion models, which require transmitting the generative models. When communication efficiency needs to be prioritized, adopting few-shot FedMP described earlier results in the least performance degradation. It is observed that with only three rounds of communication between clients and the server, where the first stage involves 30 epochs of local training and the subsequent two stages involve 60 epochs of local training each, the resulting ensemble model achieves performance comparable to baseline methods with multiple rounds of communication, demonstrating significant communication cost savings in the FL system. We provide few-shot experimental results for other FL methods in the supplementary material.

\begin{table}[ht]
\centering
\caption{Communication overhead and final accuracy (TB).}
\vspace{-4pt}
\begin{tabular}{l|c|c|c}
\hline
\textbf{Methods} & \makecell{\textbf{Communica-} \\ \textbf{tion Rounds}} & \makecell{\textbf{Communication} \\ \textbf{Cost (bytes)}} & \textbf{Accuracy} \\
\hline
FedAvg                & 65  & 36.68G   & 84.45 \\
FedProx               & 60  & 33.86G   & 85.84 \\
FedBN                 & 35  & 19.71G   & 81.86 \\
MOON                  & 50  & 28.21G   & 84.45 \\
FRAug                 & 55  & 163.92G  & 86.70 \\
Elastic Aggregation   & 55  & 31.04G   & 86.53 \\
FedMR                 & 45  & 25.39G   & 85.41 \\
FedMP        & 60   & 34.38G    & 88.08 \\
Few-Shot FedMP        & 3   & 1.71G    & 85.49 \\
\hline
\end{tabular}
\label{tab:communication_efficiency}
\vspace{-15pt}
\end{table}

\subsubsection{Privacy Preservation Analysis}
Since FedMP involves transmitting representations derived from client's private data, we evaluate the privacy leakage risks via a reconstruction attack. 
We assume that an attacker intercepts both the transmitted feature vectors and the associated feature extractor during communication, and possesses an auxiliary dataset drawn from a distribution similar to that of the client’s private data, which is used to train a decoder model to reconstruct images from features. 
In our experiment, we simulate this attack using an encoder-decoder framework. For each client, the encoder is the frozen feature extractor from FedMP framework, while the decoder, consisting of deconvolutional layers, is trained using MSE loss on the subset (50\%) of local images.
We assume that the attacker intercepts the set of feature vectors from a specific local dataset, which are then inputted to corresponding pre-trained decoder. The degree of privacy exposure is assessed by comparing the reconstructed and original images using three metrics:
(1) Fréchet Inception Distance (FID)\cite{heusel2017gans}, which measures the distance between two sets of images by comparing means and covariance matrices of features from a pre-trained inception network. A lower FID score indicates greater privacy leakage. 
(2) Structural Similarity Index Measure (SSIM)\cite{wang2004image}: SSIM evaluates image similarity based on luminance, contrast, and structure. An SSIM score closer to 1 indicates that the reconstructed image is highly similar to the original, suggesting potential leakage of semantic content. (3) $L_2$ distance: Following the memorization threshold proposed in \cite{carlini2023extracting}, we use the pixel-wise $L_2$ distance between reconstructed and original images as a risk indicator.
If the value is below the threshold $0.1$, the reconstructed feature is considered to pose a privacy risk.

We conduct experiments on the DR dataset, evaluating the reconstructed images against the original private data using the FID, maximum SSIM, and minimum $L_2$ distance. The final results are reported as the average of experiments on three clients. As shown in Table~\ref{tab:privacy}, we compare privacy leakage under different manifold dimensions. The results show that when features from stage 2, 3, or 4 are transmitted, privacy leakage metrics remain within a safe threshold. 
It indicates that deeper-layer features, which contain more abstract and less semantically detailed information, result in less accurate attacking reconstructions and a lower degree of privacy leakage.
This supports the conclusion that utilizing deeper feature layers in the FedMP framework provides better privacy protection while still enabling effective model optimization.

\begin{table}[ht]
\centering
\caption{Privacy leakage degree corresponding to different features used.}
\vspace{-5pt}
\begin{tabular}{l|c|c|c}
\hline
\textbf{Feature Layer} & \textbf{FID} $\uparrow$ & \textbf{Max SSIM} $\downarrow$ & \textbf{Min $L_2$ Distance $\uparrow$} \\
\hline
1st stage                & 820  & 0.7661   & 0.0801 \\
2nd stage               & 841  & 0.7429   & 0.1010 \\
3rd stage                 & 955  & 0.7218   & 0.1823 \\
4th stage                  & 1047  & 0.6992   & 0.1979 \\
\hline
\end{tabular}
\label{tab:privacy}
\end{table}

\section{Conclusions}
In this paper, we propose FedMP, a robust and broadly applicable federated learning algorithm that directly and effectively addresses the feature heterogeneity challenges. FedMP enhances the discriminative capability of local classifiers and aligns feature distributions of the same category across clients through (1) stochastic feature manifold completion (SFMC) and (2) class-prototype guided manifold alignment (cPGMA). 
Comprehensive experiments on various datasets, including real-world feature non-IID data, demonstrate that FedMP consistently outperforms existing FL methods. In addition, we provide an in-depth analysis of the privacy-preserving properties of the method. 
Furthermore, we demonstrate that FedMP can be adapted to a communication-efficient few-shot training paradigm, thereby alleviating the communication overhead in the FL system. 

\bibliography{main}

\end{document}